\documentclass[10pt,twocolumn,letterpaper]{article}

\usepackage{iccv}
\usepackage{times}
\usepackage{epsfig}
\usepackage{graphicx}
\usepackage{amsmath}
\usepackage{amssymb}
\usepackage{multirow}
\usepackage{adjustbox}
\usepackage{subcaption}
\usepackage{caption}
\usepackage{dsfont}
\usepackage[pagebackref=true,breaklinks=true,letterpaper=true,colorlinks,bookmarks=false]{hyperref}

\iccvfinalcopy % *** Uncomment this line for the final submission

\def\iccvPaperID{0000} % *** Enter the ICCV Paper ID here

\ificcvfinal\fi

\begin{document}

%%%%%%%%% TITLE
\title{Knowledge Guided Semi-Supervised Learning for Quality Assessment of User Generated Videos}

\author{Shankhanil Mitra \quad  Rajiv Soundararajan\\
Visual Information Processing Lab, \\
Indian Institute of Science, Bengaluru, India\\
{\tt\small  \{shankhanilm, rajivs\}@iisc.ac.in}}

%\author[1]{Subhadeep Roy$^\ast$}
%\author[1]{Shankhanil Mitra$^\ast$} %$^\dagger$
%\author[1]{Soma Biswas}
%\author[1]{Rajiv Soundararajan}
%\affil[1]{Indian Institute of Science, Bengaluru, India
%\affil[2]{Mercedes-Benz Research and Development India
%\authorcr {\tt\small subhadeeproy2000@gmail.com, \{shankhanilm, somabiswas, rajivs\}@iisc.ac.in}}

\maketitle
% Remove page # from the first page of camera-ready.
\ificcvfinal\thispagestyle{empty}\fi

\begin{abstract}
    Perceptual quality assessment of user generated content (UGC) videos is challenging due to the requirement of large scale human annotated videos for training. In this work, we address this challenge by first designing a self-supervised Spatio-Temporal Visual Quality Representation Learning (ST-VQRL) framework to generate robust quality aware features for videos. Then, we propose a dual-model based Semi Supervised Learning (SSL) method specifically designed for the Video Quality Assessment (SSL-VQA) task, through a novel knowledge transfer of quality predictions between the two models. Our SSL-VQA method uses the ST-VQRL backbone to produce robust performances across various VQA datasets including cross-database settings, despite being  learned with limited human annotated videos. Our model improves the state-of-the-art performance when trained only with limited data by around 10\%, and by  around 15\% when unlabelled data is also used in SSL. Source codes and checkpoints are available at 
    \url{https://github.com/Shankhanil006/SSL-VQA}.
\end{abstract}

%%%%%%%%% BODY TEXT %%%%%%%%%%%%%%%%%

\section{Introduction}

The emergence of video capturing devices such as smartphones, DSLRs, and GoPro has led to millions of users uploading or accessing videos via various sharing platforms such as YouTube, Instagram, Facebook and so on. This necessitates the quality assessment (QA) of videos to monitor and control the user experience. However, a reference video is often not available for user generated content (UGC), motivating the study of no reference (NR) video QA (VQA). Further, the videos also suffer from complex camera captured distortions which makes the task of NR VQA extremely challenging. 

The recent decade has seen significant progress in NR VQA, based on classical or handcrafted features \cite{vbliind,vcornia, friquee, videval,rapique} and deep learning based approaches \cite{qa_in_the_wild,mdtvsfa,fastVQA, rirnet,hierarchical_nrvqa}. The deep learning based approaches particularly require training on large amount of labelled data, which is cumbersome and expensive to acquire. This leads to us to the question of how we can design NR VQA models which can be trained with very limited labelled training data, yet achieve excellent generalisation performance on multiple datasets in terms of correlation with human perception. 

Our focus in this work is on designing semi-supervised NR VQA method with limited labelled along with unlabelled data. Since UGC videos have diverse quality characteristics, we believe that pretraining a robust video quality feature backbone is extremely important to transfer knowledge during semi-supervised learning. With this motivation, we approach the problem using a combination of contrastive self-supervised pretraining followed by semi-supervised finetuning.
A few self-supervised contrastive learning based methods have been designed for NR VQA recently \cite{vision,CONVIQT,cspt} to learn rich video quality features. However, none of these methods yet exploit the performance benefit offered by the attention mechanism in transformer based models. One of the major challenges in training such transformer based architectures for VQA is the difficulty in training such networks end-to-end. We leverage recent literature on end-to-end training of Swin-transformers for supervised VQA \cite{fastVQA}  to overcome this difficulty in self-supervised video quality representation learning. Further, we employ a novel statistical contrastive learning loss instead of a point-wise similarity loss to make the learning more robust. Thus, in the first stage of our approach, we learn rich quality aware spatio-temporal features without requiring any human annotations. 

In the second stage of our approach, we leverage the limited number of quality labels in a semi-supervised learning (SSL) framework. While several SSL based methods on pseudo-labelling and consistency regularisation have been explored in video action recognition \cite{cmpl,TCL,video_action_ssl}, they need to adapted for the specific task of VQA. In this direction, we employ knowledge transfer between two measures of video quality evaluated on the unlabelled videos. The first measure is based only on human annotations, while the second measure uses a distance between features of the distorted video and a corpus of pristine videos along with human labels. Such knowledge transfer helps overcome the drawback of limited human annotations, while simultaneously trying to help determine a perceptually relevant distance to a corpus of pristine videos. We show that the above semi-supervised learning helps design an effective VQA method with limited labels.

We conduct several experiments on multiple cross and intra datasets to validate the performance of our proposed framework. %\textcolor{red}{Particularly, we achieve impressive cross-dataset performance although we train with very few labels}.
To summarize, our main contributions consist of:
\begin{enumerate}
    \item Self-supervised statistical contrastive learning of spatio-temporal video quality representations with a transformer based architecture.
    \item Semi-supervised learning of video quality by knowledge transfer between models based on limited human labels and feature distances to a corpus of pristine videos. 
    \item Impressive cross-database performance despite the model being trained with very few human annotated videos. 
\end{enumerate}

\section{Related Work}

\textbf{Classical Feature based VQA}. Historically, handcrafted heuristics based features have been shown to produce robust performance across various VQA datasets. Among them, VBLIINDS \cite{vbliind} and VCORNIA \cite{vcornia} learn natural scene statistics of video frames by modelling the discrete cosine transform (DCT) or 3D-DCT. In recent years, TLVQM \cite{tlvqm} has shown considerable improvement in VQA performance by modelling temporal low complexity features with spatial high complexity features. VIDEVAL \cite{videval} is an ensemble of various handcrafted features designed to capture diverse quality attributes in a video. Nevertheless, authentically distorted videos in UGC have mixed distortions, which are very hard to model using the above statistical methods.

\textbf{Supervised pretraining based VQA}. Existing deep learning methods mostly regress fixed quality aware features against human opinion scores due to the computational complexity of training large models. VSFA \cite{qa_in_the_wild} and MDTVSFA \cite{mdtvsfa} learn a gated recurrent unit on top of features generated by ResNet50 \cite{resnet}. PVQ \cite{patchVQ} extracts 2D and 3D pretrained features from image quality assessment (IQA) and action recognition tasks. Recently, FAST-VQA \cite{fastVQA} learns an end-to-end model by spatially fragmenting the video clips thus reducing the complexity. TCSVT-BVQA \cite{csvt_bvqa} on the other hand learns a VQA model by transferring spatial knowledge from pretrained IQA, and temporal knowledge from a pretrained action recognition model. 

\textbf{Unsupervised pretraining based VQA}. VISION \cite{vision}, and CONVIQT \cite{CONVIQT} present self-supervised learning based quality aware feature extractors. The fixed features from these self-supervised models can be further regressed against opinion scores to develop an end-to-end quality model. In our work, we first train a self-supervised quality feature extractor and use it to build an end-to-end SSL framework.

\textbf{Unsupervised VQA}. VQA methods such as STEM \cite{STEM}, VISION \cite{vision}, and NVQE \cite{NVQE} do not require any human labelled videos in their design and give reasonable quality estimates for UGC videos. Nevertheless, their performance with respect to the methods trained with human opinion scores is under par.

\textbf{Semi-Supervised Learning}. To the best of our knowledge, there exist no end-to-end SSL algorithms designed for the VQA task. SSL methods for classification can be broadly classified into pseudo-labelling, consistency regularisation, and hybrid methods. While pseudo-labelling as such is unsuitable for regression, consistency regularisation and its hybrid versions such as Mean Teacher \cite{mean_teach}, FixMatch \cite{fixmatch}, MixMatch \cite{mixmatch}, and Meta Pseudo-Label \cite{mpl} are better suited for regression tasks. In the case of QA, these algorithms can not be directly applied as augmentations for image/video classification are quality variant. 

We remark that SSL has not been explored much even in the IQA literature. While Conde \etal \cite{semi_iqa_conformer} study SSL methods for full reference IQA, some other methods \cite{semi_iqa_kedema, sslIQA} train an NR IQA model with a large number of labelled images and generate pseudo-labels on the unlabelled data. 
\section{Spatio-Temporal VQ Representation Learning} \label{sec:stvqrl}

\textbf{Overview}. 
First, we learn a self-supervised spatio-temporal backbone to capture Video Quality (VQ) aware features  from unlabelled videos (Sec. \ref{sec:stvqrl}). The VQ representation based feature extractor is used as a backbone in our semi-supervised model (Sec. \ref{sec:ssl_vqa}) to get robust performance despite learning on limited data. 

We embark on solving multiple key challenges in learning a 3D self-supervised representation learning for VQA. It is computationally hard and even infeasible to train a video transformer using contrastive learning with videos of high resolution. Inspired by recent works on VQA \cite{fastVQA, maxVQA}, we propose a quality invariant sampling strategy that preserves the global context and local quality of videos to overcome the computational challenges while training such models. In addition, a 3D vision transformer such as video swin transformer \cite{video_swin_t} captures both short and long duration temporal distortions such as shakiness, motion blur, and flicker in videos on account of its design. Finally, in traditional contrastive learning \cite{simclr,cmc, iic_ssl}, a point-wise similarity between the global representation of features is optimized, which ignores the local variations in video space-time. To address this problem, we propose a statistical contrastive loss where both global and local information are shared between a contrasting pair of video features.
\begin{figure}
\begin{center}
\includegraphics[width=0.9\columnwidth, keepaspectratio]{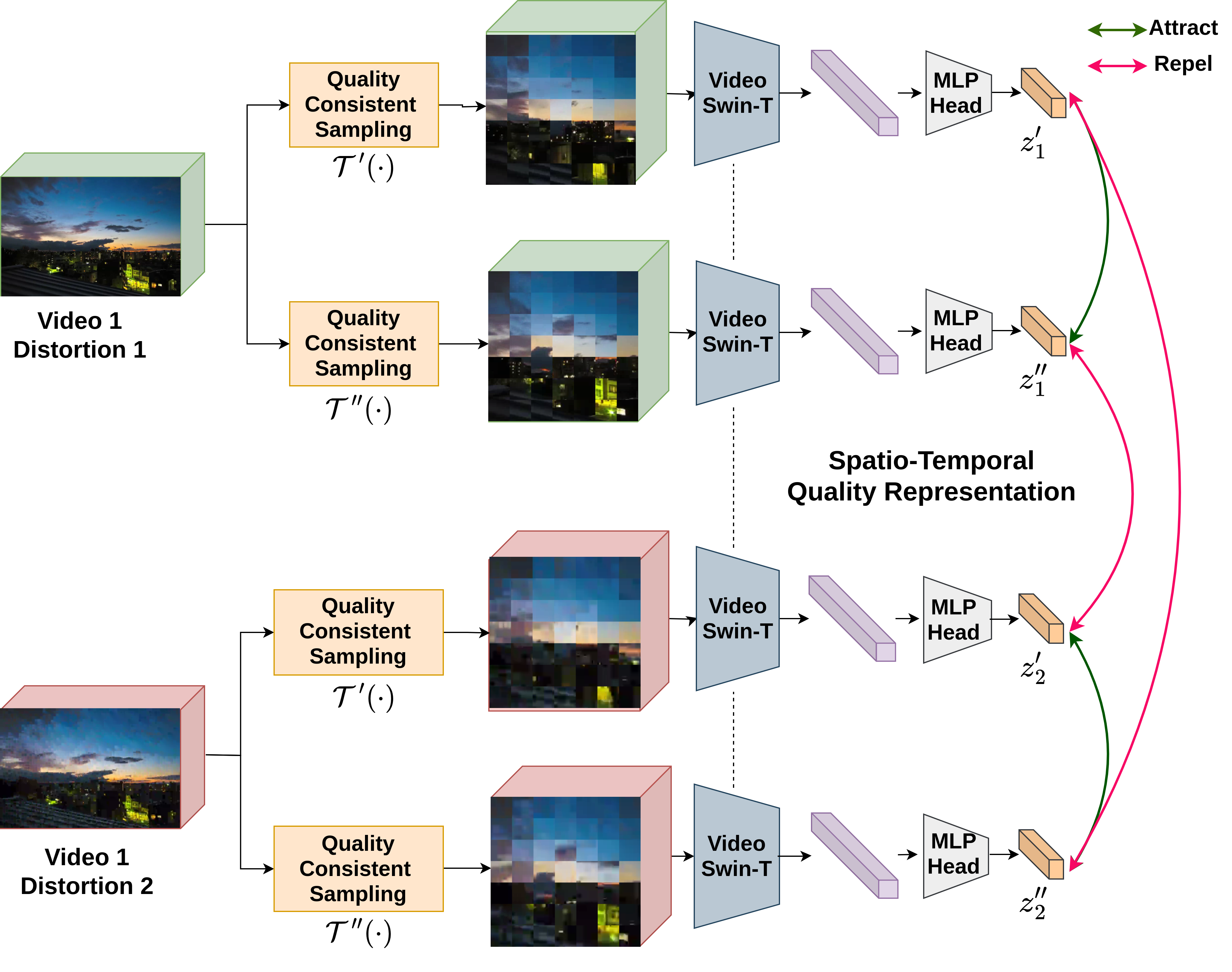}
\end{center}
\caption{Framework of  Spatio-Temporal VQ Representation Learning (\textbf{ST-VQRL}). For two distorted versions of the same video, Video 1, we sample an augmented pair of clips from each distorted version using \textbf{QCS}.  The quality aware contrastive loss is used to train the network to attract augmented clips from same video and repel clips from other distorted versions in the embedding space. }
\label{fig:stvqrl_framework}
\end{figure}

\textbf{Quality Consistent Sampling and Video Augmentation.} To capture quality-aware representations, we choose contrastive pairs of video clips from synthetically distorted UGC videos having similar content but different levels and types of distortions. We synthetically distort UGC videos similar to VISION \cite{vision} to model mixed camera captured and synthetic distortions from which quality aware features can be learned. We then employ spatial fragment sampling (crop multiple patches at original resolution and splice them together) of continuous video frames to capture both local and global video distortions in frames \cite{fastVQA}. 
As shown in literature \cite{fastVQA, maxVQA}, random fragments sampled from a video clip, share similar quality representations. We refer to the process of obtaining fragments from video clips as quality consistent sampling (\textbf{QCS}). 

Let $\mathcal{T}'(\cdot)$ and $\mathcal{T}''(\cdot)$ denote the generators of two instances of random QCS. We apply $\mathcal{T}'(\cdot)$ and $\mathcal{T}''(\cdot)$  on the same clip as shown in Figure \ref{fig:stvqrl_framework} thus generating augmented clips having similar local and global quality.  
Augmented versions of a clip constitute a positive pair. Fragments sampled from different distorted versions of the same video clip using $\mathcal{T}'(\cdot)$ and $\mathcal{T}''(\cdot)$ constitute a negative pair.

\textbf{Statistical Contrastive Loss for Representation Learning}. We capture spatio-temporal representations from a video clip using a Video Swin-T \cite{video_swin_t} backbone given as $f_{\theta}(\cdot)$ with model parameters $\theta$.  Consider a set of $K$ video clips $\{ V_1, V_2, \ldots, V_K\}$ of the same scene content with different distortions. Let, $\mathcal{T}'(V_i)$ and $\mathcal{T}''(V_i)$ denote the pair of augmented clips of every video $V_i$. Let $z_i' = f_{\theta}(\mathcal{T}'(V_i))$, and $z_i'' = f_{\theta}(\mathcal{T}''(V_i))$ be the feature representations of the augmented pair of clips of $V_i$, where $z$ is of dimension $N\times C$.
We propose a statistical constrastive loss between the spatio-temporal feature representation of the pairs to capture both the global description and local variations in the representations of fragments. 

Our contrastive loss minimises a distance between the augmented pair of representations $z_i'$, and $z_i''$ and maximises the distance between $z_i'$, and $z_k'$, where $k\neq j$ and $k\in\{1,2,\ldots,K\}$.  In particular, we are inspired by the work in NIQE \cite{niqe}, where it is shown that a statistical distance between image features is relevant to perceptual quality.  We treat the $N\times C$ feature vector as a set of $N$ spatio-temporal samples of dimension $C$ drawn from a multivariate Gaussian (MVG) model. Let the MVG model parameters be $(\mu', \Sigma')$, and $(\mu'', \Sigma'')$ for feature representations $z'$, and $z''$. Thus, we obtain a quality aware distance between any $z'$, and $z''$ as

\begin{equation}
    d(z',z'') = \sqrt{(\mu' -\mu'')^T \left(\frac{\Sigma' +\Sigma''}{2} \right)^{-1} (\mu' - \mu'')}. 
    \label{niqe_distance}
\end{equation}

Therefore, the quality aware contrastive loss with $\mathcal{T}'(V_i), i\in\{1,2,\ldots,K\}$ taken as an anchor view is $ \mathcal{L}' = \frac{1}{K} \sum_{i=1}^K l_i' $, where 

\begin{equation}
     l_i' = - \log \frac{\exp(- d(z'_i,z''_i)/ \tau)}{\sum_{j=1}^K\exp(- d(z'_i,z''_j)/ \tau)}.
     \label{contrastive_loss}
\end{equation}
Similarly, taking $\mathcal{T}''(V_i), i\in\{1,2,\ldots,K\}$ as anchor, we obtain a loss $\mathcal{L}''$, and the overall loss is given as,
\begin{equation}
    \mathcal{L}_c = \mathcal{L}' + \mathcal{L}''.
    \label{contrastive_objective}
\end{equation}

\begin{figure*}
\begin{center}
%\fbox{\rule{0pt}{2in} \rule{.9\linewidth}{0pt}}
\includegraphics[width=0.9\textwidth,keepaspectratio]{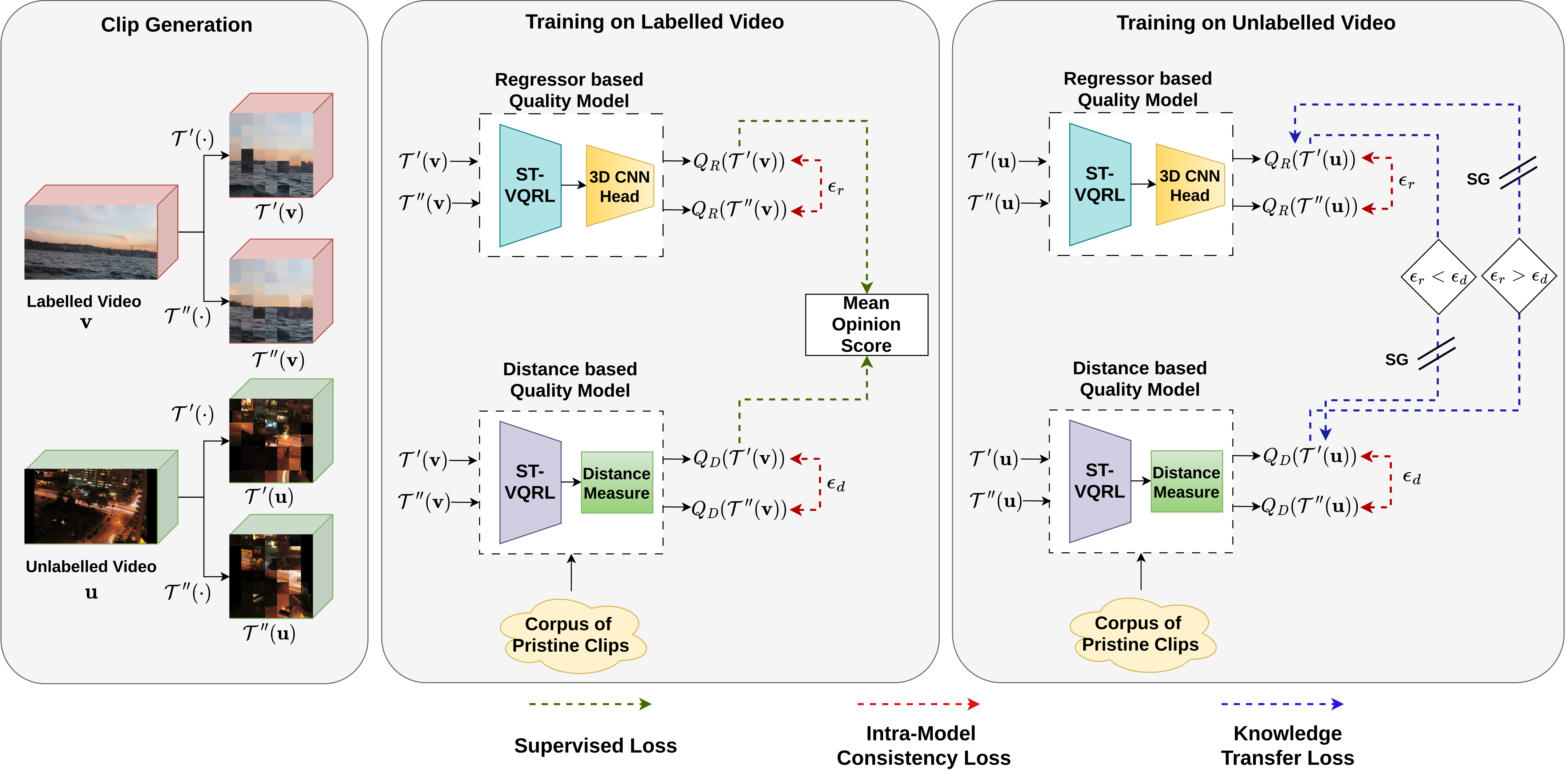}
\end{center}
\caption{Overview of Semi-Supervised Learning for VQA (\textbf{SSL-VQA}) method. Every batch consists of labelled $(\textbf{v} \in V_v)$ and unlabelled $(\textbf{u} \in U)$ samples. First, we generate an augmented pair of clips from $\textbf{v}$ and $\textbf{u}$ using two instances of quality consistent sampling (QCS), $\mathcal{T}'(\cdot)$ and $\mathcal{T}''(\cdot)$. For labelled video $\textbf{v}$, SSL-VQA optimises the supervised and intra-model consistency loss. For unlabelled video $\textbf{u}$, SSL-VQA enforces intra-model consistency loss and knowledge transfer loss. Based on the consistency criteria $(\epsilon_r > \epsilon_d)$, SSL-VQA transfers knowledge from one model to another.}
\label{fig:sslvqa_framework}
\end{figure*}

\section{Knowledge Transfer based SSL-VQA} \label{sec:ssl_vqa}
Given a set of labelled UGC videos $V= \{ (v_1,y_1), \cdots (v_{N_l},y_{N_l}) \}$ (annotated with human opinion scores), and a set of unlabelled videos $U = \{u_1, \cdots u_{N_u} \}$, our proposed approach learns quality assessment of UGC videos by utilizing both sets. 
As shown in Figure \ref{fig:sslvqa_framework}, we design a dual-model learning setup,  where one model directly maps the video features to a scalar video quality score while the other model maps the distance between the representations of a distorted video and corpus of pristine videos to video quality. The two models differ in the use of a corpus of pristine videos to predict quality. While the use of distance to a corpus imposes more structure in the quality prediction, it may also limit the quality modelling capability. Our goal is to transfer quality aware knowledge learned by the individual models to each other. While the backbone in both the models is initialised using pretrained ST-VQRL (with parameter $\theta$), we update their parameter separately as $\theta'$, and $\theta''$ respectively during finetuning.

\textbf{Regressor based Quality Model}. We attach a regressor head on top of the spatio-temporal feature encoder viz. ST-VQRL and train this model end-to-end as shown in Figure \ref{fig:sslvqa_framework}. Let $g_\phi(\cdot)$ be a non-linear regressor head with parameter $\phi$ applied on top of self-supervised ST-VQRL feature extractor $f_{\theta'}(\cdot)$ to predict a scalar quality estimate of videos. $g_\phi(\cdot)$ comprises of two 3d convolutional layers with filter size $1\times 1\times 1$ to preserve the local quality characteristics of the generated features of $f_{\theta'}(\cdot)$. Therefore, the predicted quality for any video $x \in (V_v \cup U)$, where $V_v = \{v_i\}_{i=1}^{N_l}$ is given as,
\begin{equation*}
    Q_R(x) = g_\phi(f_{\theta'}(x)).
\end{equation*}

\textbf{Distance based Quality Model}. The second model $f_{\theta''}(\cdot)$ described in Figure \ref{fig:sslvqa_framework} is necessarily a feature encoder like ST-VQRL. $f_{\theta''}(\cdot)$  measures the distance between the feature representation of a corpus of pristine videos and any distorted video $x \in (V_v \cup U)$. Let, $z^x = f_{\theta''}(x)$ denote the feature representation of $x$. Similarly, for $N_p$ pristine or clean videos, we get feature embedding $z^r$ using $f_{\theta''}(\cdot)$. Thereafter, we fit a multivariate Gaussian (MVG) model on the corpus of pristine feature set $z^r$ to get $(\mu^r, \Sigma^r)$. Let the model parameters for the distorted video representation $z^x$ be $(\mu^x, \Sigma^x)$. The quality estimate of video $x$ is given as $Q_D(x) = \exp(-d(z^r,z^x)/\tau)$, where
\begin{equation*}
    d(z^r,z^x) = \sqrt{(\mu^r -\mu^x)^T \left(\frac{\Sigma^r +\Sigma^x}{2} \right)^{-1} (\mu^r - \mu^x)}.
\end{equation*}
We predict the overall quality of a video during inference stage as $(Q_R(y)+ Q_D(y))/2$, where $y$ is any test video.

\subsection{Supervised Learning Loss}
We train both the models with the labelled videos using the ground truth opinion scores. Both models are trained separately on a mini-batch of size $B_l$ by minimising the batch-wise Pearson's linear correlation coefficient (PLCC) between predicted quality and ground truth opinion scores. Note that this loss is differentiable and allows for back-propagation. Let $\mathbf{v} = \{v_i\}_{i=1}^{B_l}$ and $\mathbf{y} = \{y_i\}_{i=1}^{B_l}$, where $\{(v_i,y_i)\}_{i=1}^{B_l} \in V$. Thus, the supervised loss is formulated as
\begin{equation}
    \mathcal{L}_s = \mathcal{L}_{plcc}(Q_R(\mathcal{T}'(\mathbf{v})), \mathbf{y}) + \mathcal{L}_{plcc}(Q_D(\mathcal{T}'(\mathbf{v})), \mathbf{y}),
\label{mos_loss}
\end{equation}
where $\mathcal{L}_{plcc}(a,b) = \frac{1 - PLCC(a,b)}{2}$ and $\mathcal{T}'(\mathbf{v}) = \{\mathcal{T}'(v_i) \}_{i=1}^{B_l}$. Since learning end-to-end on the original video resolution is computationally intensive, we apply QCS as described in Section \ref{sec:stvqrl}.

\subsection{Intra-Model Consistency Loss}

The goal of the intra-model consistency loss is to enable consistent quality predictions for the augmented video clips obtained through QCS. Let the two quality consistent augmented clips for a mini-batch of videos $\mathbf{x} = \{ x_i \}_{i=1}^{B}$, where $\{ x_i \}_{i=1}^{B} \in (V_v \cup U)$ be $\mathcal{T}'(\mathbf{x})$, and $\mathcal{T}''(\mathbf{x})$. Note that $ B =B_l + B_u$, where $B_u$ is the mini-batch length of unlabelled videos. The intra-model consistency loss is given as
%\begin{equation}
%\begin{split}
\begin{align}
    \mathcal{L}_c =& \mathcal{L}_{plcc}(Q_R(\mathcal{T}'(\mathbf{x})), Q_R(\mathcal{T}''(\mathbf{x})))) \nonumber\\
                  &+ \mathcal{L}_{plcc}(Q_D(\mathcal{T}'(\mathbf{x})), Q_D(\mathcal{T}''(\mathbf{x})))).
    \label{cons_loss}
\end{align}
%\end{split}    
%\end{equation}

\begin{table*}
\centering
\begin{tabular}{cc|cccc|cccc}
\hline
\multicolumn{2}{c|}{Setting}                      & \multicolumn{4}{c|}{Intra Test Database}                            & \multicolumn{4}{c}{Cross Database}                                \\ \cline{1-10} 
\multicolumn{2}{c|}{Test Dataset}                                       & \multicolumn{2}{c|}{LSVQ\textsubscript{test}}      & \multicolumn{2}{c|}{LSVQ\textsubscript{1080p}} & \multicolumn{2}{c|}{KoNVid-1K}     & \multicolumn{2}{c}{LIVE VQC} \\ \hline
\multicolumn{1}{c|}{Method}   & Model Type                    & SROCC & \multicolumn{1}{c|}{PLCC}  & SROCC          & PLCC          & SROCC & \multicolumn{1}{c|}{PLCC}  & SROCC         & PLCC         \\ \hline
\multicolumn{1}{c|}{VBLIIND }  & \multirow{3}{*}{\begin{tabular}[c]{@{}c@{}}Classical\\ Features\end{tabular}}    & 0.473 & \multicolumn{1}{c|}{0.456} & 0.382          & 0.430         & 0.545 & \multicolumn{1}{c|}{0.539} & 0.398         & 0.434        \\
\multicolumn{1}{c|}{TLVQM }    &                               & 0.599 & \multicolumn{1}{c|}{0.582} & 0.441          & 0.473         & 0.592 & \multicolumn{1}{c|}{0.597} & 0.531         & 0.551        \\
\multicolumn{1}{c|}{VIDEVAL }  &                               & 0.607 & \multicolumn{1}{c|}{0.597} & 0.491          & 0.547         & 0.545 & \multicolumn{1}{c|}{0.543} & 0.416         & 0.459        \\ \hline
\multicolumn{1}{c|}{VSFA}     & \multirow{2}{*}{\begin{tabular}[c]{@{}c@{}}\\Supervised\\ Pretraining\end{tabular}}   & 0.663 & \multicolumn{1}{c|}{0.645} & 0.536          & 0.543         & 0.664 & \multicolumn{1}{c|}{0.669} & 0.651         & 0.668        \\
\multicolumn{1}{c|}{FAST-VQA } &                               & 0.682 & \multicolumn{1}{c|}{0.677} & 0.552          & 0.558         & 0.679 & \multicolumn{1}{c|}{0.666} & 0.652         & 0.672        \\ 
\multicolumn{1}{c|}{TCSVT-BVQA } &                               & 0.687 & \multicolumn{1}{c|}{0.679} & 0.449          & 0.465         & 0.682 & \multicolumn{1}{c|}{0.680} & 0.665         & 0.682        \\ \hline
\multicolumn{1}{c|}{VISION }   & \multirow{4}{*}{\begin{tabular}[c]{@{}c@{}}\\ Self-\\Supervised\\ Pretraining\end{tabular}} & 0.523 & \multicolumn{1}{c|}{0.478} & 0.427          & 0.446         & 0.606 & \multicolumn{1}{c|}{0.612} & 0.615         & 0.636        \\
\multicolumn{1}{c|}{CONVIQT }  &                               & 0.636 & \multicolumn{1}{c|}{0.624} & 0.468          & 0.464         & 0.662 & \multicolumn{1}{c|}{0.673} & 0.600         & 0.627        \\ \cline{1-1} \cline{3-10}
\multicolumn{1}{c|}{\begin{tabular}[c]{@{}c@{}}$\textrm{SSL-VQA}^{-}$\\ w/o consistency loss\end{tabular}}  &               & 0.704   & \multicolumn{1}{c|}{0.693} &  0.562         & 0.577        &  0.715 & \multicolumn{1}{c|}{0.713} & 0.669      & 0.678        \\ \cline{1-1} \cline{3-10}
\multicolumn{1}{c|}{$\textbf{SSL-VQA}^{-}$}      &   & \textbf{0.719} & \multicolumn{1}{c|}{\textbf{0.717}} & \textbf{0.587}      & \textbf{0.601}         & \textbf{0.736} & \multicolumn{1}{c|}{\textbf{0.735}} & \textbf{0.683}         & \textbf{0.688}        \\ \hline
\end{tabular}
\caption{Performance analysis of $\textrm{SSL-VQA}^{-}$ with ST-VQRL backbone compared against other popular VQA methods with classical features, supervised pretrained, and self-supervised pretrained backbones when trained only with limited human annotated videos \textbf{without any unlabelled data}.}
\label{supervised_table}
\end{table*}

\subsection{Knowledge Transfer based Loss}
The goal of the knowledge transfer loss is to utilise both the models effectively for SSL. If a consistency between the prediction of both the models on unlabelled data is enforced, an erroneous prediction of one model may drive the other model to wrong knowledge. So the challenge here is to transfer knowledge from one model to another only when its prediction is reliable. We hypothesize that we can rely on the prediction of the model if it is stable with respect to augmented versions of a sample. To determine model stability with respect to a batch, we evaluate the intra-model consistency for a mini-batch of unlabelled samples $\textbf{u} = \{ u_i \}_{i=1}^{B_u}$, where each $ u_i \in U$. The model whose consistency loss between the augmented pair of videos in Equation (\ref{cons_loss}) is less, is considered more stable. Let $\epsilon_r$ denote the regressor based quality model's consistency error and $\epsilon_d$ denote the distance based quality model's consistency error. Then,

\begin{equation*}
\begin{split}
    \epsilon_r &= \mathcal{L}_{plcc}(Q_R(\mathcal{T}'(\mathbf{u})), Q_R(\mathcal{T}''(\mathbf{u})))\\
    \epsilon_d &= \mathcal{L}_{plcc}(Q_D(\mathcal{T}'(\mathbf{u})), Q_D(\mathcal{T}''(\mathbf{u}))) .
\end{split}
\end{equation*}
The prediction of the more stable model is used as a pseudo-label for the other model. The knowledge transferable loss function for a mini-batch is given as

\begin{equation}
\begin{split}
    \mathcal{L}_u &= m \mathcal{L}_{plcc}(Q_R(\mathcal{T}'(\mathbf{u})), sg(Q_D(\mathcal{T}'(\mathbf{u})))))\\
                  &+ (1 -m)\mathcal{L}_{plcc}(sg(Q_R(\mathcal{T}'(\mathbf{u}))), Q_D(\mathcal{T}'(\mathbf{u})))),
\end{split}
\label{stab_loss}
\end{equation}
where  $m=\mathds{1}(\epsilon_r > \epsilon_d)$ is the indicator \textit{mask} and $sg(.)$ denotes the stop gradient operation. This ensures that the stable model provides a pseudo-label or guidance for the other model.

\begin{table*}
\centering
\begin{tabular}{c|cccc|cccc}
\hline
\multicolumn{1}{c|}{Setting} & \multicolumn{4}{c|}{Intra Test Database}                              & \multicolumn{4}{c}{Cross Database}                               \\ \hline 
\multicolumn{1}{c|}{Test Database} & \multicolumn{2}{c|}{LSVQ\textsubscript{test}}  & \multicolumn{2}{c|}{LSVQ\textsubscript{1080p}} & \multicolumn{2}{c|}{KoNVid-1K}    & \multicolumn{2}{c}{LIVE VQC} \\ \hline 
Method                & SROCC & \multicolumn{1}{c|}{PLCC} & SROCC  & PLCC        & SROCC & \multicolumn{1}{c|}{PLCC} & SROCC         & PLCC         \\ \hline
Mean Teacher                    & 0.716 & \multicolumn{1}{c|}{0.703}& 0.594  & 0.603       & 0.716 & \multicolumn{1}{c|}{0.715}&  0.679        & 0.681             \\
Meta PseudoLabel                   & 0.714 & \multicolumn{1}{c|}{0.713}& 0.586  & 0.587       & 0.719 & \multicolumn{1}{c|}{0.716}&  0.676        &  0.673            \\
FixMatch                   & 0.722 & \multicolumn{1}{c|}{0.725}& 0.582  & 0.596       & 0.727 & \multicolumn{1}{c|}{0.732}& 0.685         & 0.687             \\ \hline
\textbf{SSL-VQA}                   & \textbf{0.731} & \multicolumn{1}{c|}{\textbf{0.736}}&  \textbf{0.616} & \textbf{0.645}       & \textbf{0.765} & \multicolumn{1}{c|}{\textbf{0.770}} &  \textbf{0.711}  & \textbf{0.734}      \\ \hline
\end{tabular}
\caption{Performance comparison of SSL-VQA with other SSL benchmarks on intra and inter database test settings. All methods are initialised with \textbf{ST-VQRL backbone} for fair comparison and are trained \textbf{with both labelled and unlabelled samples}.} 
\label{semisupervised_table}
\end{table*}

The overall loss to train our SSL-VQA model end-to-end is a combination of the supervised loss, intra-model consistency loss, and knowledge transferable loss as

\begin{equation}
    \mathcal{L} = \mathcal{L}_s + \lambda_c  \mathcal{L}_c + \lambda_u \mathcal{L}_u,
    \label{overall_loss}
\end{equation}
where $\lambda_c$, $\lambda_u$ are hyper parameters to balance the loss terms.

\section{Experiments} \label{sec:experiments}
 In this section, we describe the implementation details, experimental setup, and comparisons with other methods.

\subsection{Implementation Details of ST-VQRL}

\textbf{Data Generation.} As mentioned in Section \ref{sec:stvqrl}, we learn our self-supervised ST-VQRL model on a set of synthetically distorted UGC videos. We randomly sample 200 videos out of 28056 training videos of LIVE-FB Large-Scale Social Video Quality (LSVQ) \cite{patchVQ} database. LSVQ database videos have unique scenes with camera captured distortions, so we augment each of the 200 videos with 12 different synthetic distortion types and levels in the same
manner as in VISION \cite{vision}. In the statistical contrastive loss and distance based quality model, we use a set of 60 pristine videos from LIVE-VQA \cite{live_sd1}, LIVE Mobile \cite{mobile1}, CSIQ VQD \cite{csiq}, EPFL-PoLiMI \cite{epfl1} and ECCV-EVVQ databases \cite{ecvq_evvq2}.

\textbf{Training Details}.
We encode the distorted video sequence using a Video Swin-T \cite{video_swin_t} architecture modified with gated relative positional bias \cite{fastVQA} to take into account the discontinuity in a sampled clip due to QCS. QCS is applied by dividing each of the 32 continuous video frames into a $7\times 7$ grid, sampling $32\times 32$ patches from each grid and stitching them together maintaining temporal consistency. The patches within each grid are extracted from the same location for every distorted version of a scene, thus the distorted clips are content consistent with respect to sampling. We train ST-VQRL using AdamW \cite{adamw} with a learning rate of $10^{-4}$ and a weight decay of $0.05$ for 30 epochs. The temperature co-efficient $\tau$ mentioned in Equation (\ref{contrastive_loss}) is  $10$.

\subsection{Experimental Setup} %\label{sec:experiments}
We conduct two types of experiments to evaluate VQA under limited labelled data. In the first experiment (\textbf{Experimental Setting 1}) in Table \ref{supervised_table}, we define $\textrm{SSL-VQA}^{-}$ as our model learnt with only limited labelled samples and without using any unlabelled data. Thus, we only use the losses in Equations (\ref{mos_loss}) and (\ref{cons_loss}) to understand how the framework can learn with just the limited labelled data available for training. In the second experiment (\textbf{Experimental Setting 2}) in Table \ref{semisupervised_table}, we perform SSL by also utilizing the unlabelled data. In both the cases we train the model for 30 epochs using AdamW \cite{adamw} with a learning rate of  $10^{-4}$ and a weight decay of $0.05$. $\lambda_c$, and $\lambda_u$ are chosen to be 1 based on training loss convergence.%We describe the training and testing datasets for both experiments below. 

\textbf{Training Database.} LSVQ \cite{patchVQ} has an official training set of 28056 videos. We extract 2000 videos from it randomly. Out of the 2000 videos, we only use around 500 videos or $1.78\%$ of the training set with human opinion scores and the remaining 1500 unlabelled videos.

\textbf{Evaluation Database.} We employ two different test settings. In the first setting, we test on the official test database $\textrm{LSVQ}_{\textrm{test}}$, 
 containing 7400 videos of varying resolution between 240p and 720p, and $\textrm{LSVQ}_{\textrm{1080p}}$ containing around 3600 videos of 1080p for intra database performance evaluation. We further test the robustness of SSL-VQA in cross database settings. Particularly, we test on KoNVid-1K \cite{konvid}, and LIVE VQC \cite{livevqc}, each comprising of 1200 and 585 camera captured authentically distorted videos with varying resolutions. We use the Spearman Rank-Order Correlation Coefficient (SROCC), and Pearson Linear Correlation Coefficient (PLCC) as performance measures. In both Table \ref{supervised_table} and \ref{semisupervised_table}, we report the median performance over 3 random choices of 500 labelled and 1500 unlabelled videos out of the 2000 chosen videos. 

\begin{table*}
\centering
\begin{tabular}{c|cc|cc|cc|cc}
\hline
                 & \multicolumn{2}{c|}{KoNVid-1K} & \multicolumn{2}{c|}{LIVE VQC} & \multicolumn{2}{c|}{LIVE QCOMM} & \multicolumn{2}{c}{YouTube-UGC} \\ \cline{2-9} 
Method           & SROCC          & PLCC          & SROCC         & PLCC          & SROCC          & PLCC           & SROCC          & PLCC           \\ \hline
Mean Teacher     & 0.783          & 0.786         & 0.702         & 0.693         & 0.722          & 0.724          & 0.725          & 0.730          \\
Meta PseudoLabel & 0.792          & 0.795         & 0.705         & 0.710         & 0.734          & 0.735          & 0.729          & 0.719               \\
FixMatch         & 0.798          & 0.799         & 0.716         & 0.729         & 0.740          & 0.730          & 0.734          & 0.730          \\ \hline
\textbf{SSL-VQA}             & \textbf{0.826}          & \textbf{0.828}         & \textbf{0.733}         & \textbf{0.743}         & \textbf{0.747}          & \textbf{0.751}          & \textbf{0.750}          & \textbf{0.757}          \\ \hline
\end{tabular}
\caption{Performance on KoNViD, LIVE-VQC, LIVE Qualcomm and YouTube-UGC when SSL-VQA is finetuned on $20\%$ of annotated videos from each database. We provide comparison with other SSL benchmarks which are also finetuned on $20\%$ of labelled videos for each of these databases.}
\label{finetune}
\end{table*}

\subsection{Experimental Setting 1 Analysis}
We compare $\textrm{SSL-VQA}^{-}$ with three popular categories of NR VQA models under Experimental Setting 1. Among the models based on classical or heuristic based feature designs, we compare with Video BLIINDS \cite{vbliind}, TLVQM \cite{tlvqm}, and VIDEVAL \cite{videval}. Among recent deep VQA models, we compare with the state-of-the-art FAST-VQA model \cite{fastVQA} and the popular VSFA \cite{qa_in_the_wild} and TCSVT-BVQA \cite{csvt_bvqa} methods. Finally, we also compare with recent self-supervised feature learning models such as VISION \cite{vision} and CONVIQT \cite{CONVIQT}. The comparison with the supervised pre-trained FAST-VQA model is particularly interesting since the experiment reveals how our self-supervised ST-VQRL backbone of $\textrm{SSL-VQA}^{-}$ enables learning with limited labelled data. 

In Table \ref{supervised_table}, we see that our method outperforms both classical feature based and recent deep learning methods. This increment in performance can be attributed to the use of a robust quality aware feature backbone viz. ST-VQRL. The ST-VQRL encoder not only provides robust performance in the limited data regime but also is independent of any supervision like FAST-VQA, VSFA, and TCSVT-BVQA.

\subsection{Experimental Setting 2 Analysis} \label{sec:ssl_benchmark}

Since there exist no direct end-to-end semi-supervised VQA models in the literature for comparison, we adapt popular semi-supervised methods for the VQA task. Semi-supervised approaches can be broadly divided into pseudo-labelling and consistency regularisation. Since direct pseudo-labelling is not applicable for regression tasks, we rely upon consistency regularisation based methods such as MeanTeacher \cite{mean_teach} and FixMatch \cite{fixmatch}. We also modify the meta learning based SSL method, Meta Pseudo-Label \cite{mpl} for comparison. We use our self-supervised feature encoder ST-VQRL as the backbone for fair comparison with SSL-VQA approaches. 

%\subsection{Performance Analysis of SSL}
In Table \ref{semisupervised_table}, we provide a quantitative comparison between SSL-VQA and other SSL methods modified for VQA. In both intra-database and cross database settings, we see a considerable improvement of SSL-VQA over other methods. Thus a smart knowledge transfer between the two quality models enriches our SSL framework leading to superior performance. 

\section{Ablation Studies}

\textbf{Finetuning on UGC Data:}
In Section \ref{sec:experiments}, we show SSL-VQA's robustness across intra and inter database test settings. Now we evaluate SSL-VQA by finetuning it for specific VQA tasks. In general, we adapt our model on four smaller VQA datasets viz. KoNVid-1k \cite{konvid}, LIVE VQC \cite{livevqc}, YouTube-UGC \cite{youtube_ugc}, and LIVE Qualcomm \cite{liveqcomm}. KoNVid-1K and LIVE VQC predominantly have videos captured in the wild with cameras. YouTube-UGC comprises of videos from various domains such as real-world, animation, and gaming, spanning a spatial resolution from 240p to 4K. LIVE Qualcomm on other hand, comprises of authentic videos with specific categories of distortions such as shakiness, stabilisation, and so on. In this experiment, we randomly sample \textbf{20\%} of the videos with labels in each of the UGC databases above for finetuning and use another non-overlapping 20\% for testing. We report the median performance across 10 such splits in Table \ref{finetune}. We observe that such fine-tuning with limited labelled data substantially improves the performance. 

\textbf{Impact of Statistical Contrastive Loss:}
In Section \ref{sec:stvqrl}, we noted that the use of a statistical measure between spatio-temporal video features in Equation (\ref{niqe_distance}) is more relevant to perceptual quality. To validate our hypothesis, we show in Table \ref{backbone_variation}, the superiority of our model optimised with Equation (\ref{contrastive_objective}) over the cosine similarity loss as in generic contrastive learning. We also provide the performance when using a supervised Video Swin-T backbone \cite{video_swin_t} pretrained for action recognition over our self-supervised backbone. We infer that ST-VQRL learned from scratch, specifically to capture quality aware features gives better performance than supervised pretrained Video Swin-T. Moreover, ST-VQRL learned using a statistical distance measure captures quality representations better. 

\begin{table}
\begin{center}
\begin{tabular}{c|c|c}\hline
Backbone                                                               & KoNVid-1K & LIVE VQC \\ \hline
%No Weights                                                             &           &          \\\hline
\begin{tabular}[c]{@{}c@{}}Pre-trained \\ Video Swin-T\end{tabular}          & 0.728     & 0.686         \\\hline\hline
\begin{tabular}[c]{@{}c@{}}ST-VQRL w/ \\ Similarity loss\end{tabular}  & 0.733     & 0.684    \\ \hline
\begin{tabular}[c]{@{}c@{}}\textbf{ST-VQRL} w/ \\ \textbf{Statistical loss}\end{tabular} & \textbf{0.765}     & \textbf{0.711}     \\ \hline
\end{tabular}
\caption{SROCC performance analysis of different backbones on SSL-VQA performance.}
\label{backbone_variation}
\end{center}
\end{table}

\textbf{Role of Intra-model Consistency and Knowledge Transfer:}
SSL-VQA model is optimised using an objective function comprising of supervised loss, intra-model consistency loss, and  knowledge transfer loss. When we evaluate the model without the consistency loss, note that there is also no \textit{mask} in Equation (\ref{stab_loss}) and unrestricted knowledge transfer happens between the two models. Without the knowledge transfer loss, the unlabelled data is only used to impose intra-model consistency. In Table \ref{loss_variation}, we see that the absence of either loss hinders the learning performance showing the benefit of our contributions in SSL for VQA. 

\begin{table}
\begin{center}
\begin{tabular}{c|c|c}
\hline
Approach                  & KoNVid-1K & LIVE VQC \\ \hline
\begin{tabular}[c]{@{}c@{}}SSL-VQA \\w/o  consistency  loss\end{tabular}  & 0.756          & 0.697         \\\hline
\begin{tabular}[c]{@{}c@{}}SSL-VQA \\w/o  knowledge  loss\end{tabular} & 0.742          & 0.686         \\ \hline
\textbf{SSL-VQA}                       & \textbf{0.765}          & \textbf{0.711}         \\ \hline
\end{tabular}
\caption{SROCC performance analysis of different unsupervised constraints in Equation (\ref{cons_loss}) and (\ref{stab_loss}) respectively.} 
\label{loss_variation}
\end{center}
\end{table}

\begin{figure}[ht]
   \centering
   \begin{subfigure}[b]{0.48\columnwidth}
     \centering
     \includegraphics[width=\textwidth,keepaspectratio]{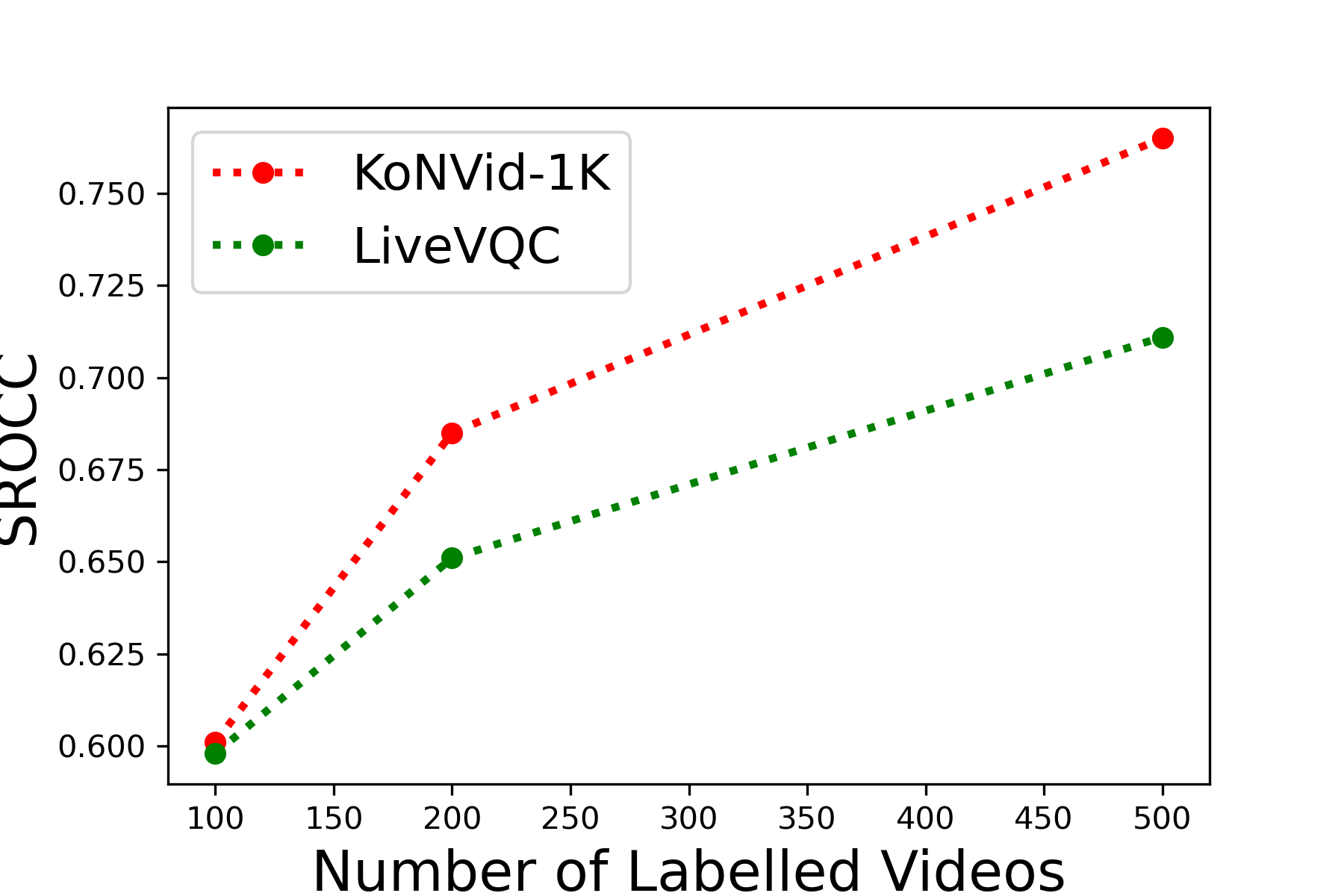}
     \caption{}
     \label{fig:labelled_kv}
   \end{subfigure}
   \hfill
   \begin{subfigure}[b]{0.48\columnwidth}
     \centering
     \includegraphics[width=\columnwidth,keepaspectratio]{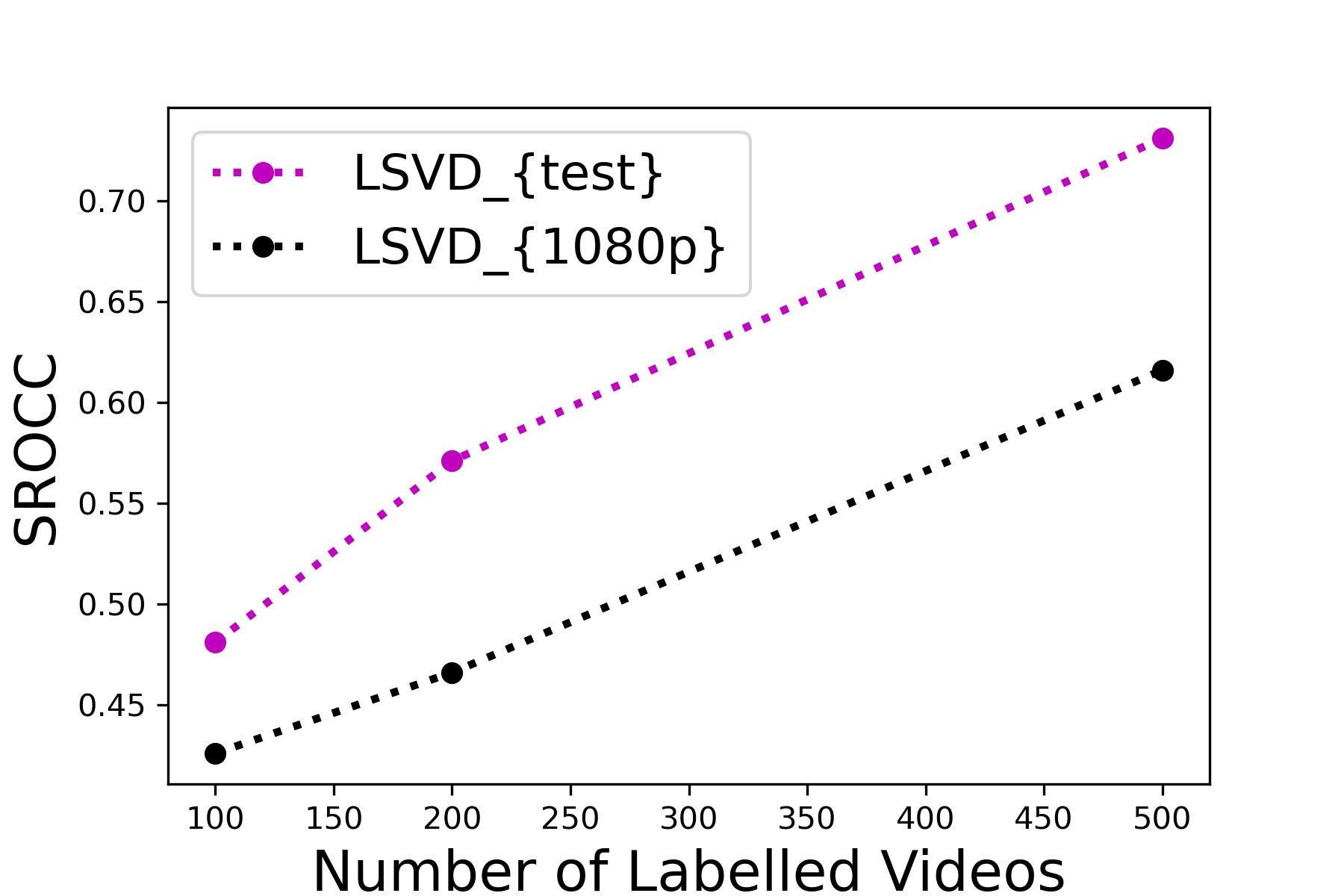}
     \caption{}
     \label{fig:labelled_lsvd}
   \end{subfigure}
   \hfill
      \begin{subfigure}[b]{0.48\columnwidth}
     \centering
     \includegraphics[width=\columnwidth,keepaspectratio]{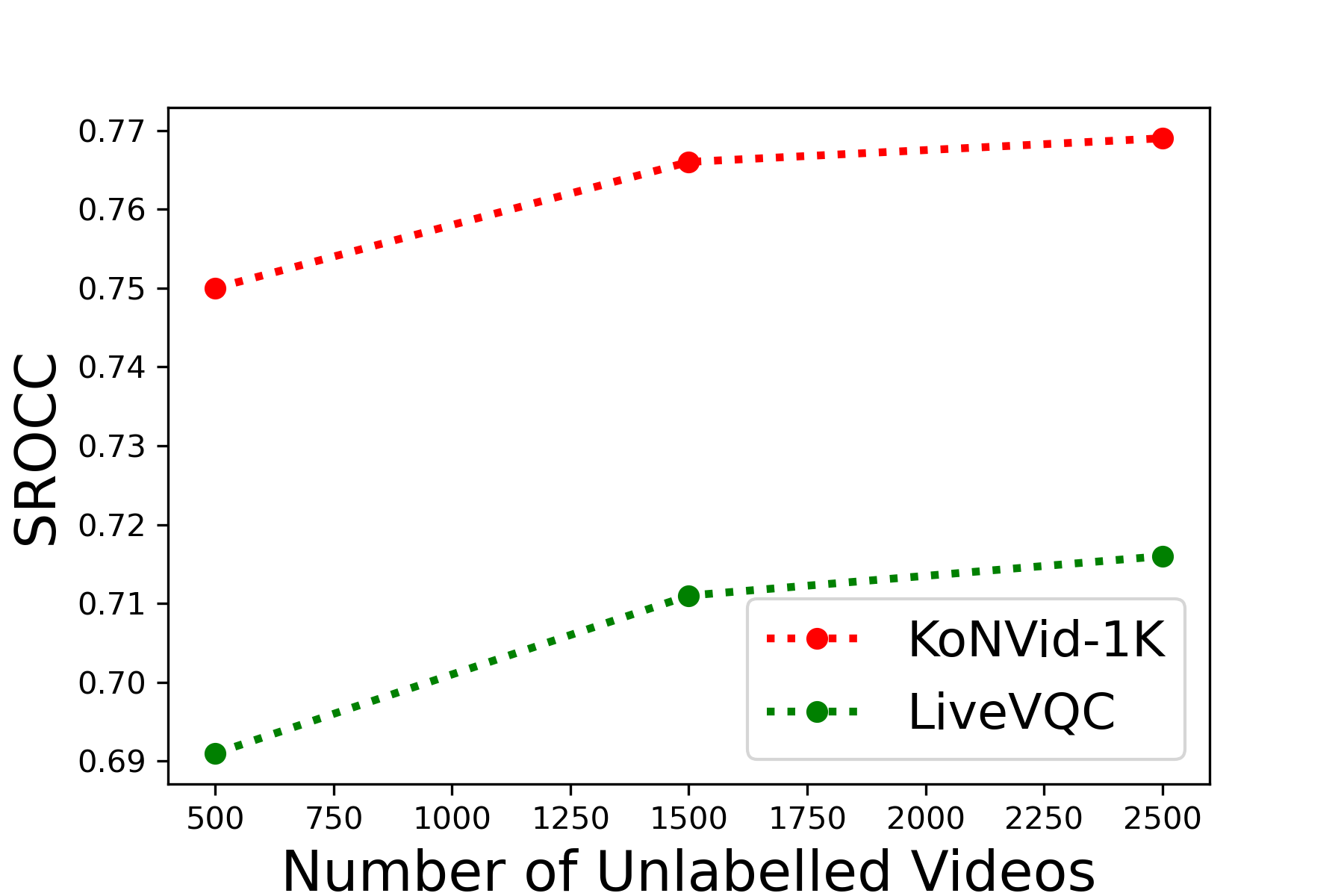}
     \caption{}
     \label{fig:unlabelled_kv}
   \end{subfigure}
   \hfill
      \begin{subfigure}[b]{0.48\columnwidth}
     \centering
     \includegraphics[width=\columnwidth,keepaspectratio]{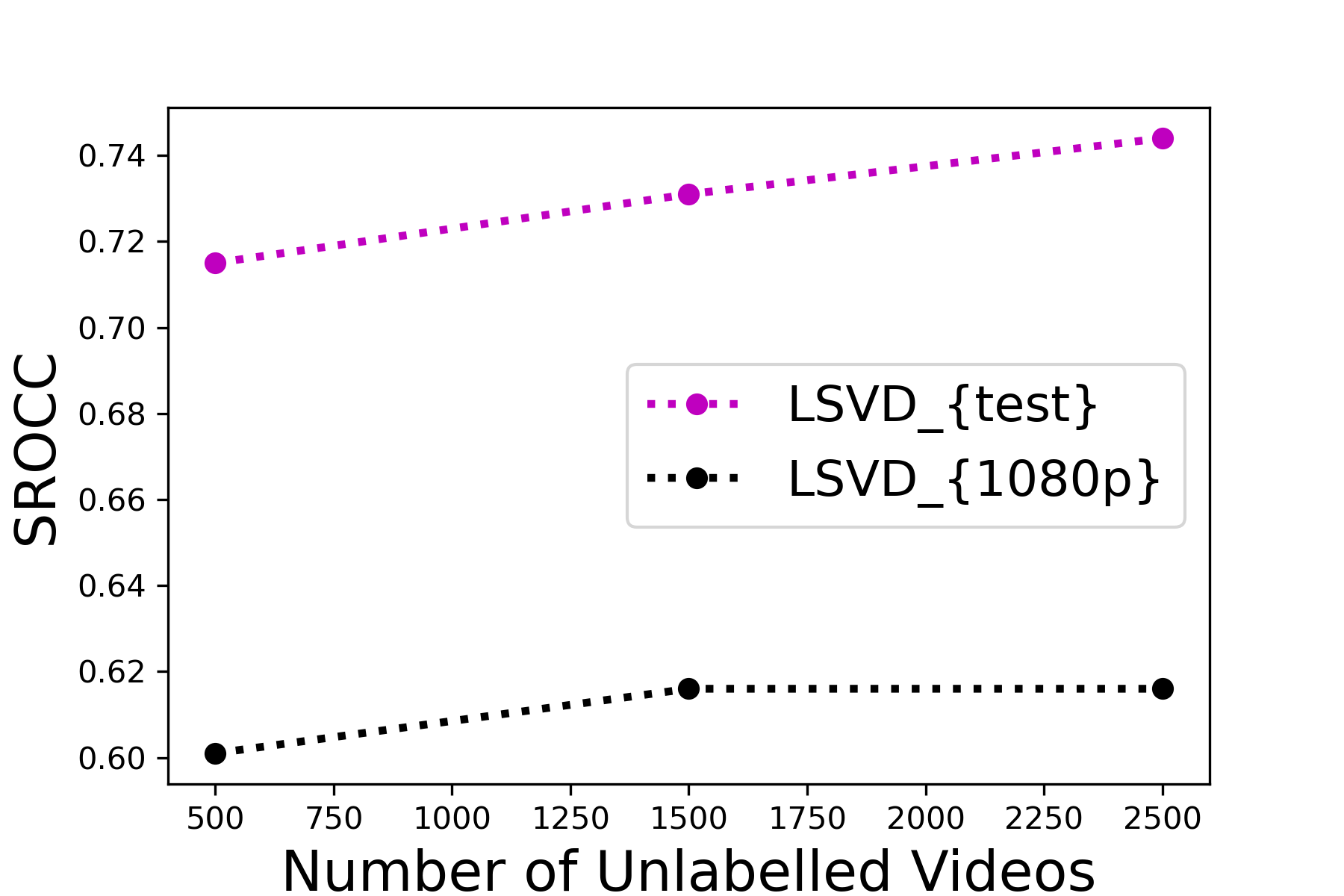}
     \caption{}
     \label{fig:unlabelled_lsvd}
   \end{subfigure}
%   \hfill
   
   \label{fig:labelled_unlabelled}
    \caption{(a), and (b) correspond to SSL-VQA performance on KoNVid-1K, LIVE-VQC, LSVQ\textsubscript{test}, and LSVQ\textsubscript{1080p} datasets when the number of labelled videos for training ranges from 100-500. In (c), and (d), we show the performance for 500 labelled videos with different amounts of unlabelled videos.}
    \label{label_variation}
\end{figure}

\textbf{Impact of Number of Labelled and Unlabelled Videos:}
As mentioned in Section \ref{sec:experiments},, we train SSL-VQA on 500 labelled and 1500 unlabelled videos from the LSVQ \cite{patchVQ} database. Here, we present an analysis of our model's performance with varying numbers of labelled and unlabelled videos. In Figure \ref{fig:labelled_kv}, and \ref{fig:labelled_lsvd}, we present the performance of SSL-VQA trained on with labelled videos varying between 100-500 keeping the number of unlabelled videos fixed. We see a steady increase in performance as the number of labelled videos increases. Similarly, in Figure \ref{fig:unlabelled_kv}, and \ref{fig:unlabelled_lsvd}, we train SSL-VQA using 500 labelled and 500-2500 unlabelled videos. We see that when the number of unlabelled videos is considerably high (more than 1500), the model's prediction nearly saturates.

\begin{table*}[]
\begin{adjustbox}{max width=\textwidth}
\begin{tabular}{c|c|c|c|c|c|c|c|c|c}
\hline
                       & VBLIIND & TLVQM   & VIDEVAL & VSFA    & FAST-VQA & TCSVT-BVQA & VISION  & CONVIQT & $\textrm{SSL-VQA}^{-}$ \\ \hline
VBLIIND                & -       & 0 0 0 0 & 0 0 0 0 & 0 0 0 0 & 0 0 0 0  & 0 0 0 0    & 0 0 0 0 & 0 0 0 0 & 0 0 0 0                \\ \hline
TLVQM                  & 1 1 1 1 & -       & 0 0 1 1 & 0 0 0 0 & 0 0 0 0  & 0 1 0 0    & 1 1 0 0 & 0 0 0 0 & 0 0 0 0                \\ \hline
VIDEVAL                & 1 1 1 1 & 1 1 0 0 & -       & 0 0 0 0 & 0 0 0 0  & 0 1 0 0    & 1 1 0 0 & 0 1 0 0 & 0 0 0 0                \\ \hline
VSFA                   & 1 1 1 1 & 1 1 1 1 & 1 1 1 1 & -       & 0 0 0 1  & 0 1 0 0    & 1 1 1 1 & 1 1 0 1 & 0 0 0 0                \\ \hline
FAST-VQA               & 1 1 1 1 & 1 1 1 1 & 1 1 1 1 & 1 1 1 0 & -        & 0 1 0 0    & 1 1 1 1 & 1 1 1 1 & 0 0 0 0                \\ \hline
TCSVT-BVQA             & 1 1 1 1 & 1 0 1 1 & 1 0 1 1 & 1 0 1 1 & 1 0 1 1  & -          & 1 1 1 1 & 1 0 1 1 & 0 0 0 0                \\ \hline
VISION                 & 1 1 1 1 & 0 0 1 1 & 0 0 1 1 & 0 0 0 0 & 0 0 0 0  & 0 0 0 0    & -       & 0 0 0 1 & 0 0 0 0                \\ \hline
CONVIQT                & 1 1 1 1 & 1 1 1 1 & 1 0 1 1 & 0 0 1 0 & 0 0 0 0  & 0 1 0 0    & 1 1 1 0 & -       & 0 0 0 0                \\ \hline
$\textrm{SSL-VQA}^{-}$ & 1 1 1 1 & 1 1 1 1 & 1 1 1 1 & 1 1 1 1 & 1 1 1 1  & 1 1 1 1    & 1 1 1 1 & 1 1 1 1 & -                      \\ \hline
\end{tabular}
\end{adjustbox}
\caption{Results of one-sided Wilcoxon Rank Sum Test performed between the SROCC values of the other VQA algorithms and $\textrm{SSL-VQA}^{-}$. Each entry in the table consists of a codeword with 4 symbols corresponding to the testing on LSVQ\textsubscript{test}, LSVQ\textsubscript{1080p}, KoNVid-1K, and LIVE VQC databases in that order. A code value of $``1"$ indicates that the VQA model in the row
is statistically superior to the VQA model in the column. While a value of $``0"$ indicates row model is inferior to the column
model and  $``-"$ indicates a statistically similar performance.}
\label{sig_fewshot}
\end{table*}

\section{Conclusion}
We presented a novel SSL approach with a robust quality aware feature encoder ST-VQRL. Through extensive experiments, we showed that SSL-VQA achieves higher performance than existing state-of-the art VQA methods even when learned with very few human annotated videos. We also benchmarked SSL methods for VQA and showed the superiority of our framework on the use of unlabelled videos. As our model works on video fragments similar to FAST-VQA \cite{fastVQA}, it is also computationally efficient. We believe that SSL-VQA can make deep learning based VQA perform robustly with limited labels.

\section*{Acknowledgment}
This work was supported in part by Department of Science
and Technology, Government of India, under grant
CRG/2020/003516.

\section*{Appendix}
\subsection*{Evaluation on full LSVQ train data}
We provide a quantitative analysis between VQA methods
and $\textrm{SSL-VQA}^{-}$ with ST-VQRL feature backbone trained
on the full annotated train set of LSVQ in Table \ref{tab:fullsupervised}. We infer
that our ST-VQRL representation achieves superior performance
compared to various benchmark methods even at full
scale supervision.
\begin{table}[]
\centering
\resizebox{\columnwidth}{!}{\begin{tabular}{c|c|c|c|c}
\hline
Method                 &  {LSVQ\textsubscript{test}} & {LSVQ\textsubscript{1080p}} &{KoNVid-1K} & {LIVE VQC}  \\ \hline
VSFA & 0.801 & 0.675 & 0.784 & 0.734 \\
PatchVQ  & 0.827 & 0.711 & 0.791 &0.770 \\
TCSVT-BVQA & 0.852 & 0.771 & 0.834 & 0.816 \\
FAST-VQA  & 0.876 & 0.779 & 0.859 & 0.823  \\
$\textbf{SSL-VQA}^{-}$  & 0.891 & 0.799 & 0.877 & 0.839  \\ \hline
%$\textbf{SSL-VQA}^{-}$  & 0.889 & 0.799 & 0.877 & 0.839  \\ \hline
\end{tabular}}
\caption{SROCC performance of various VQA methods}
\label{tab:fullsupervised}
\end{table}

\subsection*{Quality Maps Generation using SSL-VQA:}
\begin{figure*}
   \centering
   \begin{subfigure}[b]{0.45\textwidth}
     \centering
     \textbf{Video 1}\par\medskip
     \includegraphics[width=0.3\textwidth,keepaspectratio]{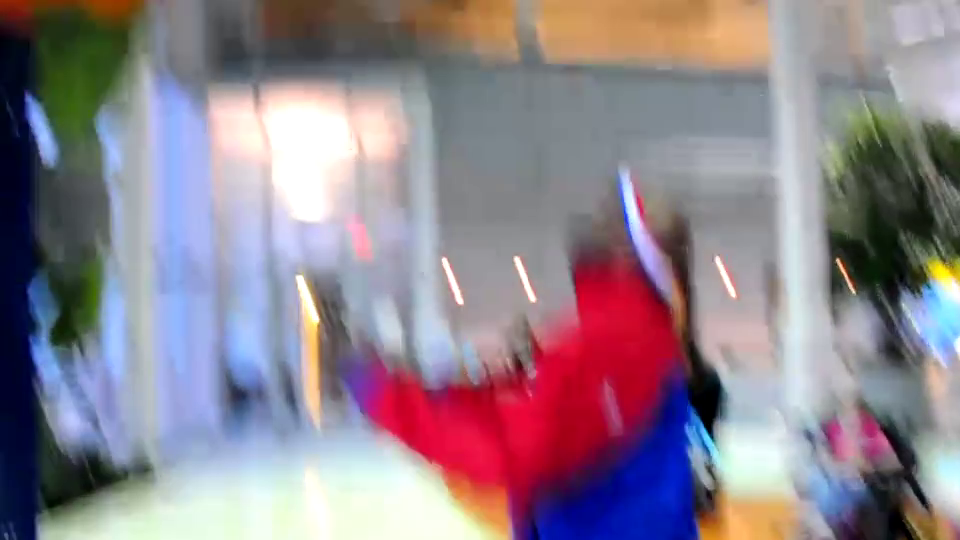} \hfill
     \includegraphics[width=0.3\textwidth,keepaspectratio]{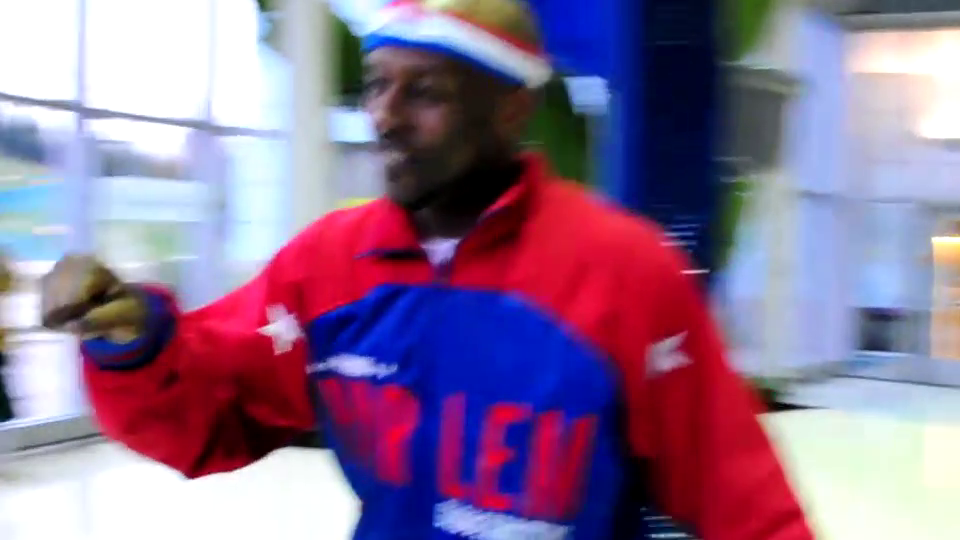} \hfill
     \includegraphics[width=0.3\textwidth,keepaspectratio]{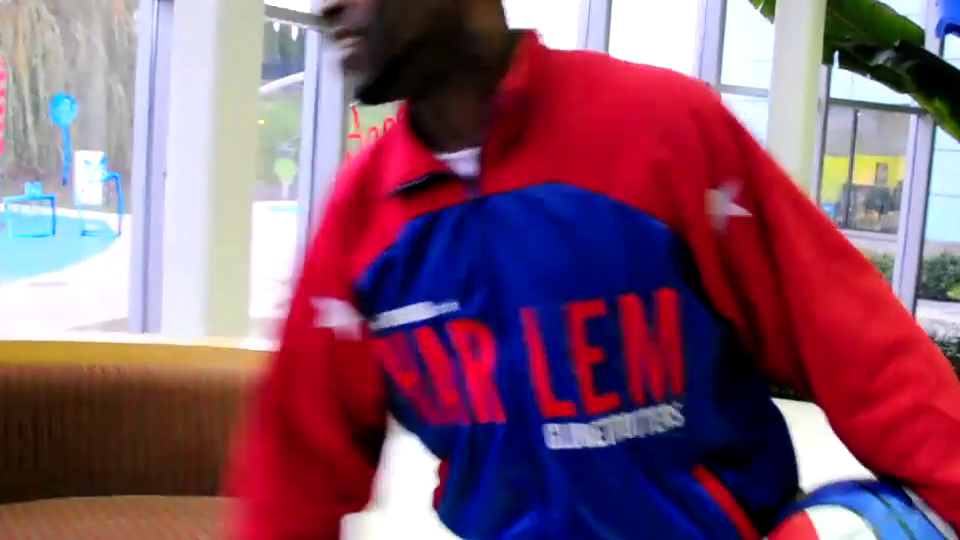} \hfill
     \caption{Original Frames from KoNVid-1K sampled at 1 fps}
     \label{konvid_org}
   \end{subfigure}
   \hfill
   \begin{subfigure}[b]{0.45\textwidth}
     \centering
     \textbf{Video 2}\par\medskip
     \includegraphics[width=0.3\textwidth,keepaspectratio]{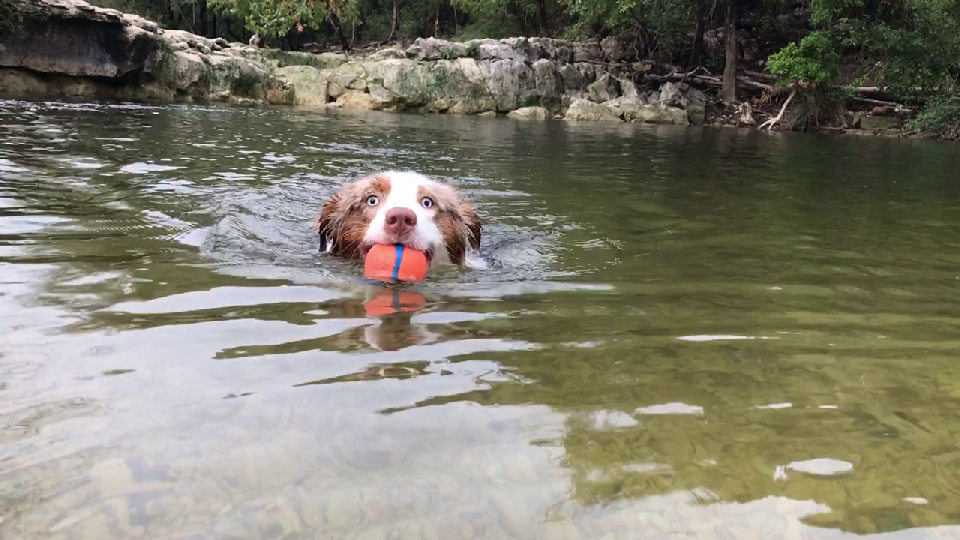} \hfill
     \includegraphics[width=0.3\textwidth,keepaspectratio]{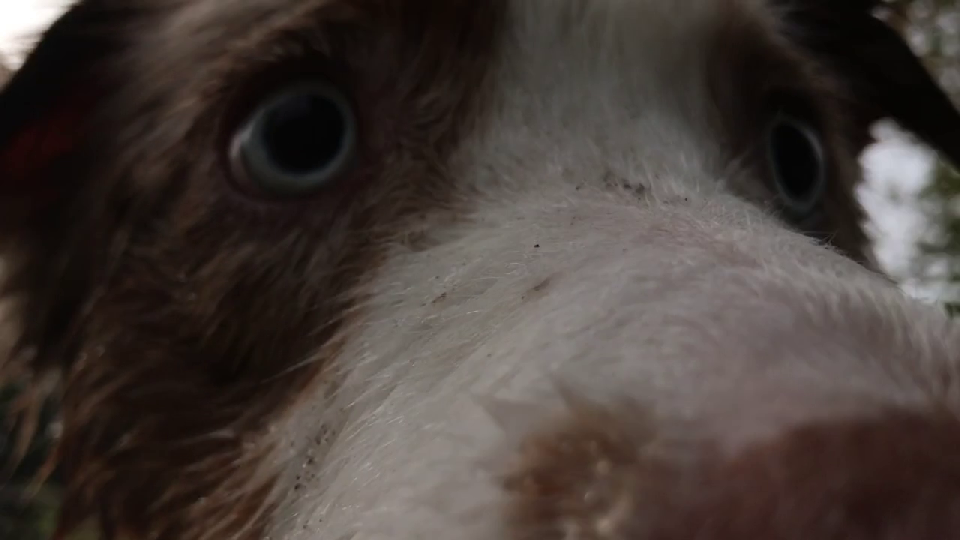}\hfill
     \includegraphics[width=0.3\textwidth,keepaspectratio]{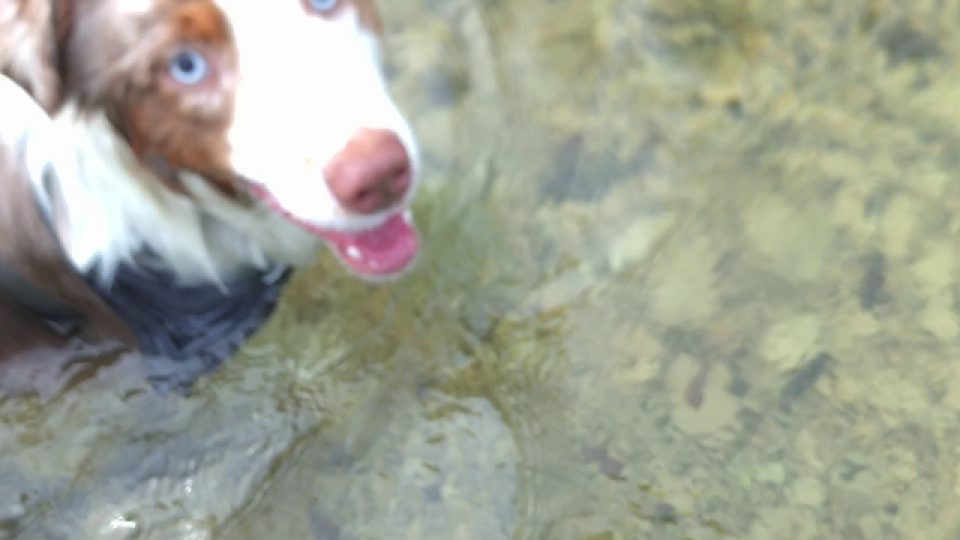}\hfill
     \caption{Original Frames from LIVE VQC sampled at 1 fps}
     \label{vqc_org}
   \end{subfigure}
   \vskip\baselineskip
      \begin{subfigure}[b]{0.45\textwidth}
     \centering
     \includegraphics[width=0.3\textwidth,keepaspectratio]{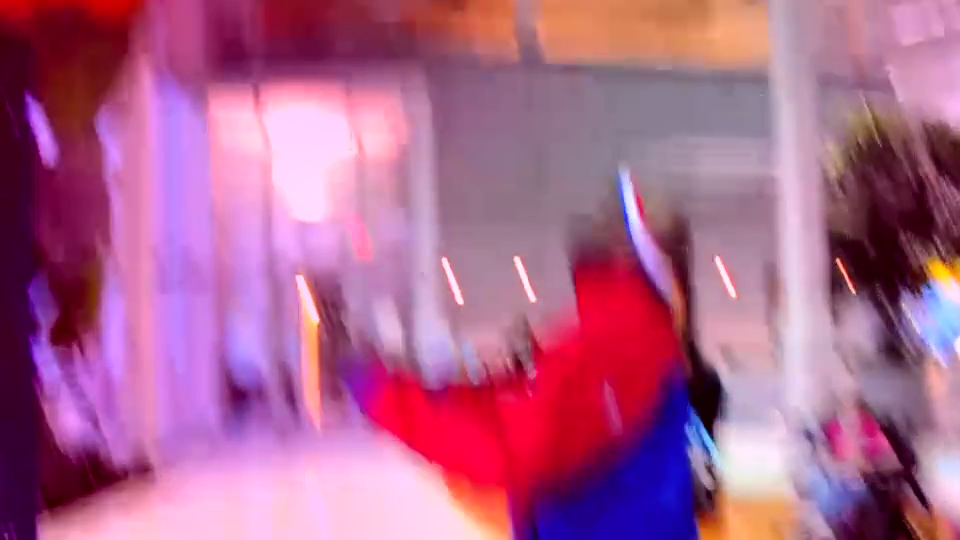} \hfill
     \includegraphics[width=0.3\textwidth,keepaspectratio]{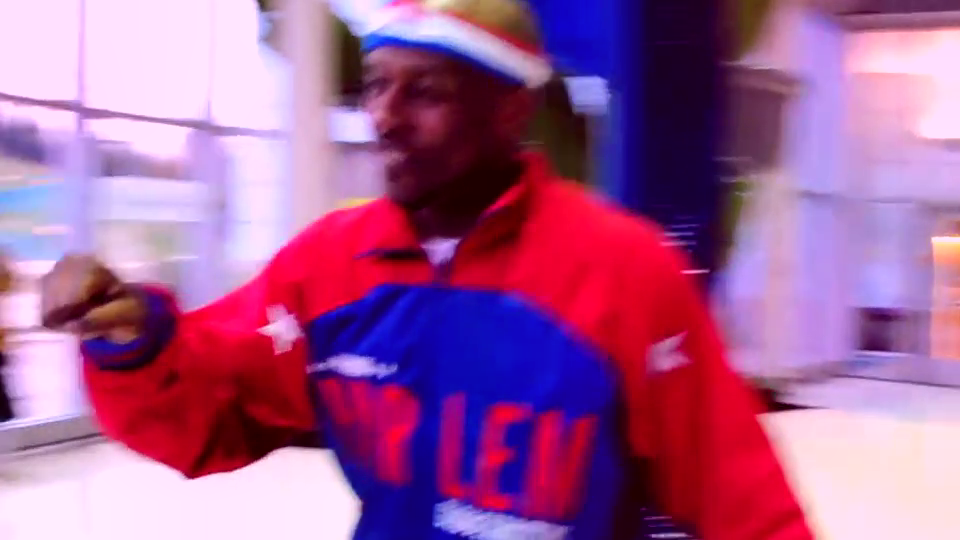} \hfill
     \includegraphics[width=0.3\textwidth,keepaspectratio]{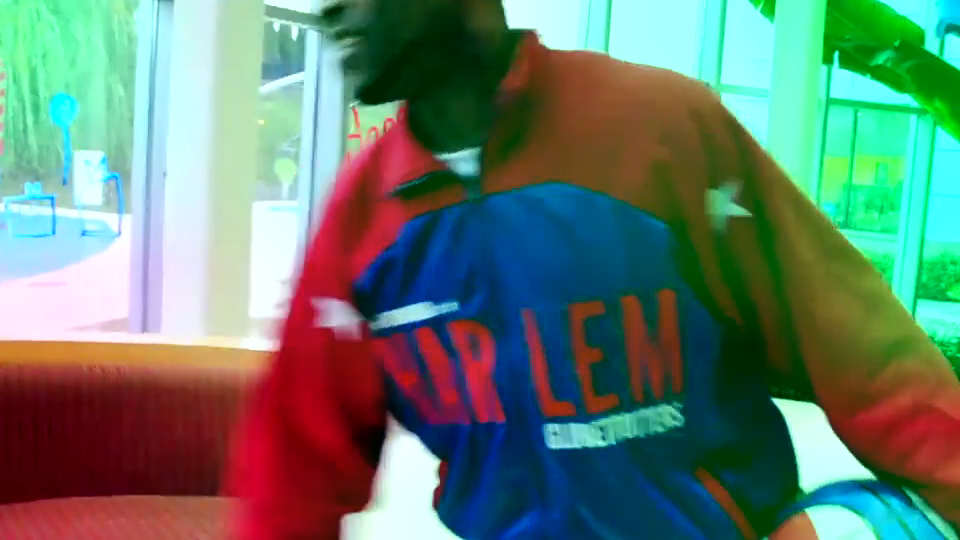} \hfill
     \caption{Generated Quality maps using our SSL-VQA}
     \label{konvid_qa}
   \end{subfigure}
   \hfill
   \begin{subfigure}[b]{0.45\textwidth}
     \centering
     \includegraphics[width=0.3\textwidth,keepaspectratio]{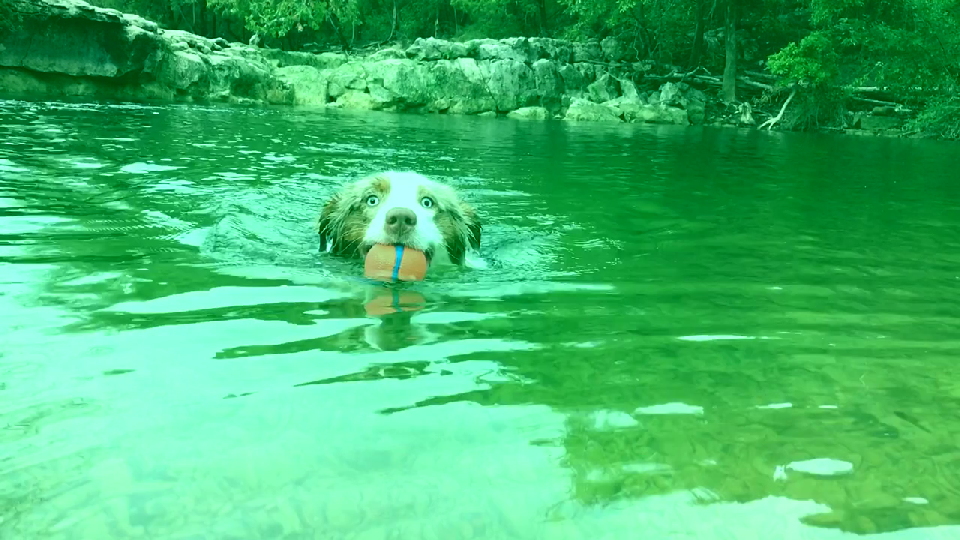} \hfill
     \includegraphics[width=0.3\textwidth,keepaspectratio]{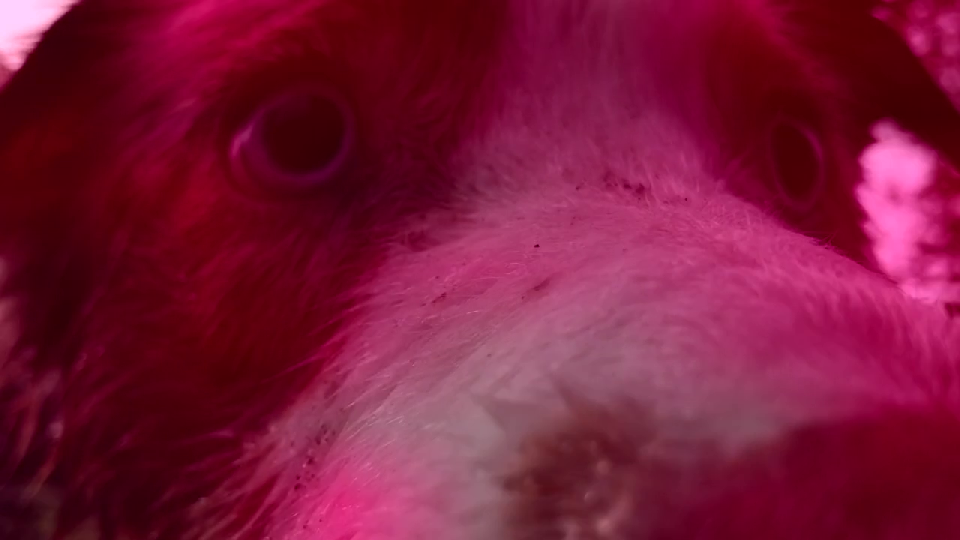}\hfill
     \includegraphics[width=0.3\textwidth,keepaspectratio]{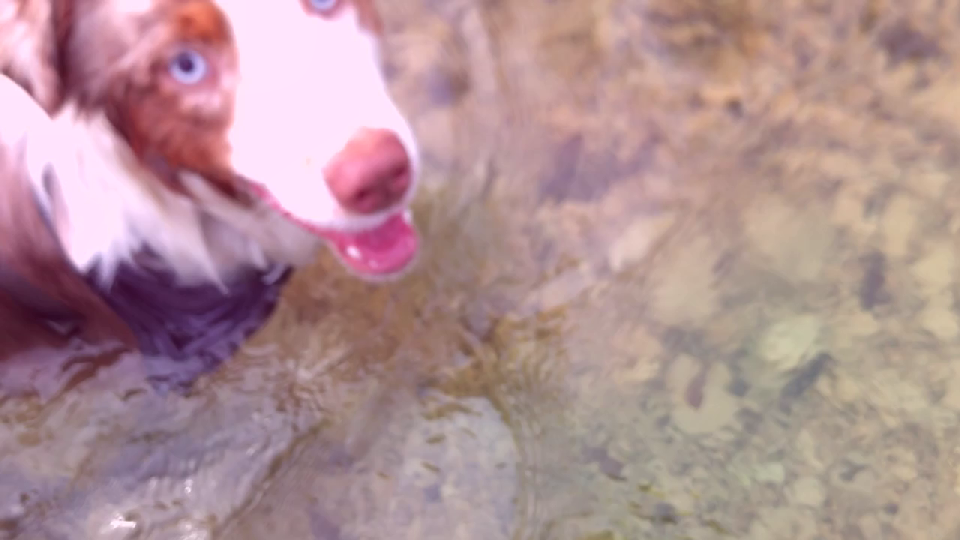}\hfill
     \caption{Generated Quality maps using our SSL-VQA}
     \label{vqc_qa}
   \end{subfigure}
   
    \caption{Original frames from KoNVid-1K (Video 1), and LIVE VQC (Video 2) sampled at 1 fps are shown in Figure \ref{konvid_org} and \ref{vqc_org}. Spatio-temporal quality maps generated by SSL-VQA re-projected to original video resolution are shown in Figure \ref{konvid_qa}, and \ref{vqc_qa}, where \textcolor{green}{green} areas correspond to high quality region, and \textcolor{red}{red} areas refer to low quality regions.}
    \label{qa_maps}
\end{figure*}

We now describe the generation of quality maps using our model and analyse them. 
The regressor head $g_\phi(\cdot)$ used in the direct regressor model predicts a quality map over the fragments which can be projected to obtain quality maps at a frame level. We generate the quality map for one video each from the KoNVid-1K and LIVE VQC datasets. In Figure \ref{qa_maps}, we show three frames from each video sampled at 1 frame per second (fps), and their respective quality maps. For Video 1 in Figure \ref{qa_maps}, the camera goes out of focus as seen from the motion blur in frames. In the quality map generated by SSL-VQA model, we that the blurry region is detected as low quality while in the third frame, the static scene is mostly detected as good quality. Similarly, for Video 2 the quality of the video suddenly degrades and again stabilises as can be seen from its quality map. Moreover, SSL-VQA correctly captures the local variation in texture, while also preserving the global context information.

\subsection*{Statistical Significance Test}
In Tables \ref{supervised_table} and \ref{semisupervised_table}, we reported the median performance of $\textrm{SSL-VQA}^{-}$ and SSL-VQA against various VQA and SSL methods in limited labelled data or in semi-supervised settings respectively over 3 random training splits. We conduct a statistical significance test to validate the superiority of our method. In particular, the non-parametric Wilcoxon Rank
Sum Test is used to compare the rank of two sets of correlation coefficients for a pair of methods across 3 splits. Similar
to \cite{twostepQA}, we consider the null hypothesis as that
the the median of one algorithm is equal to that of the other
at 95 \% significance level. The alternate hypothesis is that
the medians differ. From Tables \ref{sig_fewshot} and \ref{sig_ssl}, we observe that our $\textrm{SSL-VQA}^{-}$ and SSL-VQA outperform other methods with regard to f-test in both the experimental settings.

\begin{table}[]
\begin{adjustbox}{max width=\columnwidth}
\begin{tabular}{c|c|c|c|c}
\hline
                                                             & \begin{tabular}[c]{@{}c@{}}Mean \\ Teacher\end{tabular} & \begin{tabular}[c]{@{}c@{}}Meta \\ Pseudo\\ Label\end{tabular} & FixMatch & SSL-VQA \\ \hline
Mean Teacher                                                 & -                                                       & 0 1 0 1                                                        & 0 1 0 0  & 0 0 0 0 \\ \hline
\begin{tabular}[c]{@{}c@{}}Meta \\ Pseudo-Label\end{tabular} & 1 0 1 0                                                 & -                                                              & 0 1 0 0  & 0 0 0 0 \\ \hline
FixMatch                                                     & 1 0 1 1                                                 & 1 0 1 1                                                        & -        & 0 0 0 0 \\ \hline
SSL-VQA                                                      & 1 1 1 1                                                 & 1 1 1 1                                                        & 1 1 1 1  & -       \\ \hline
\end{tabular}
\end{adjustbox}
\caption{Results of one-sided Wilcoxon Rank Sum Test
performed between the SROCC values of the other semisupervised
algorithms and SSL-VQA. The code word has
similar representation as in Table \ref{sig_fewshot}.}
\label{sig_ssl}
\end{table}

\subsection*{Training Details}
SSL-VQA and all other benchmarking methods were trained in Python 3.8 using Pytorch 2.0 on a $3 \times 24$ GB NVIDIA RTX 3090 GPU. We consider the optimizer hyper-parameters to train SSL-VQA is similar to that of described in FAST-VQA as both uses a Video Swin-T backbone.

{\small
\bibliographystyle{ieee_fullname}
\bibliography{paper}
}
\end{document}